\documentclass[lettersize,journal]{IEEEtran}
\usepackage{amsmath,amsfonts}

\usepackage{array}
\usepackage[ruled]{algorithm2e}
\usepackage[caption=false,font=normalsize,labelfont=sf,textfont=sf]{subfig}
\usepackage{amsmath, bm}
\usepackage{textcomp}
\usepackage{stfloats}
\usepackage{url}
\usepackage{verbatim}
\usepackage{graphicx}
\usepackage{cite}
\usepackage{multirow}
\usepackage{makecell}
\usepackage{color}
\usepackage{float}
\usepackage{subfig}
\usepackage{booktabs}
\usepackage{caption}
\usepackage{tikz}
\usepackage{comment}
\usepackage{pifont}
\begin{document}

\title{Removing Multiple Hybrid Adverse Weather in Video via a Unified Model }

\author{Ye-Cong Wan, Ming-Wen Shao~\IEEEmembership{Member,~IEEE,} Yuan-Shuo Cheng, Jun Shu, and 
	Shui-Gen Wang
        \thanks{Yecong Wan, Mingwen Shao, and Yuanshuo Cheng are from China University of Petroleum (East China).}\thanks{
        Jun Shu is from Xi'an Jiaotong University.} \thanks{ Shuigen Wang is from IRay Technology Co., Ltd. }
}

\markboth{Journal of \LaTeX\ Class Files,~Vol.~14, No.~8, August~2021}%
{Shell \MakeLowercase{\textit{et al.}}: A Sample Article Using IEEEtran.cls for IEEE Journals}

\maketitle

\begin{abstract}
Videos captured under real-world adverse weather conditions typically suffer from uncertain hybrid weather artifacts with heterogeneous degradation distributions. However, existing algorithms only excel at specific single degradation distributions due to limited adaption capacity and have to deal with different weather degradations with separately trained models, thus may fail to handle real-world stochastic weather scenarios. Besides, the model training is also infeasible due to the lack of paired video data to characterize the coexistence of multiple weather.
To ameliorate the aforementioned issue, we propose a novel unified model, dubbed UniWRV, to remove multiple heterogeneous video weather degradations in an all-in-one fashion. Specifically, to tackle degenerate spatial feature heterogeneity, we propose a tailored weather prior guided module that queries exclusive priors for different instances as prompts to steer spatial feature characterization. To tackle degenerate temporal feature heterogeneity, we propose a dynamic routing aggregation module that can automatically select optimal fusion paths for different instances to dynamically integrate temporal features. 
Additionally, we managed to construct a new synthetic video dataset, termed HWVideo, for learning and benchmarking multiple hybrid adverse weather removal, which contains 15 hybrid weather conditions with a total of 1500 adverse-weather/clean paired video clips. Real-world hybrid weather videos are also collected for evaluating model generalizability.
Comprehensive experiments demonstrate that our UniWRV exhibits robust and superior adaptation capability in multiple heterogeneous degradations learning scenarios, including various generic video restoration tasks beyond weather removal.
\end{abstract}


\section{Introduction}

	\IEEEPARstart{I}{mages} and videos captured under adverse weather conditions inevitably suffer from nasty visibility and undesirable distortion due to the diverse and uncertain weather degradations under imaging. This would deliver drastical obstacles and challenges to real-world outdoor vision systems, such as autonomous driving systems \cite{liang2018deep,prakash2021multi} and surveillance systems \cite{perera2018uav}. 
To ameliorate this dilemma, various image and video adverse weather removal algorithms \cite{zamir2021multi,ren2017video,wei2017should,wang2021seeing,zhang2021learning,yang2022learning,liu2018erase} have been proposed to restore corrupted information and enhance visual presentation.

\begin{figure}
	\begin{center}
		\includegraphics[width=\linewidth]{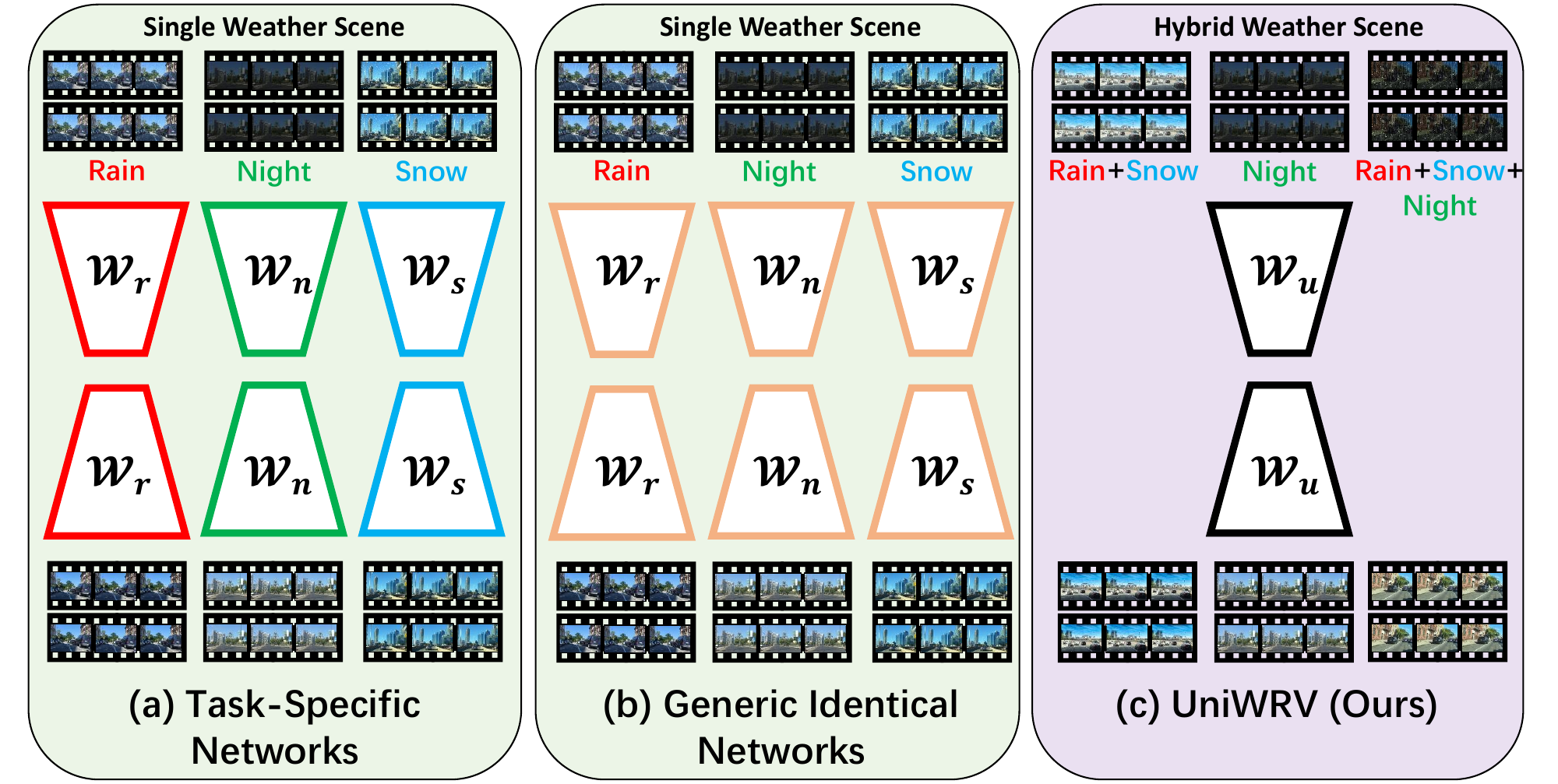}
	\end{center}
	\vspace{-7pt}
	\caption{ Overview of video adverse weather removal frameworks. (a) Separate networks designed for specific weather; (b) generic networks but with task-specific weights; (c) our proposed UniWRV framework. In contrast to existing approaches that aim to tackle different weather types with different model instances, UniWRV can handle multiple hybrid adverse weather videos with heterogeneous degradation distributions via a unified model, thus enjoying better flexibility and practicality in realistic applications.}
	\label{figure1}
	\vspace{-15pt}
\end{figure}

Unlike image restoration, video adverse weather removal methods not only exploit spatial domain information within a single frame but also probe temporal correlations of adjacent frames. Traditional video weather removal methods \cite{garg2006photorealistic,ren2017video,wei2017should} focus on modeling prior of weather effects variation in video frames to recover potential clean background. However, restricted by empirical observation and fixed representation paradigms, these approaches typically suffer from relatively inferior generalizability. Recently, learning-based approaches have attracted considerable attention and achieved promising performance.
Specifically, one category of attempts is devoted to devising weather-specific video restoration algorithms \cite{wang2021seeing,zhang2021learning,yang2022learning,liu2018erase,li2021online} (Fig. \ref{figure1}(a)) for given weather types, which are typically unable to popularize to other weather conditions. The other category of studies aims to tackle multiple degradations via identical architecture \cite{liang2022recurrent,liang2022vrt,huang2022neural,wang2019edvr} (Fig. \ref{figure1}(b)). Nevertheless, these methods are required to be trained separately and load corresponding pre-trained weights for handling different weather degradations. This would make the weather removal process troublesome and inefficient, impeding their further application on real-world systems, where diverse and uncertain multiple weather will be encountered. Despite several efforts have been made for all-in-one single image restoration \cite{chen2022learning,li2020all,li2022all,valanarasu2022transweather,ozdenizci2023restoring,zhu2023learning,park2023all,zhang2023ingredient,yang2023visual,potlapalli2023promptir,ma2023prores}, the temporal correlation among consecutive frames has remained untapped. The pioneering ViWS-Net \cite{yang2023video} is the most recent work to address video multiple-type weather removal via weather messenger and adversarial learning.
However, these approaches ignore that real-world adverse conditions often suffer from superimposed
multiple degradations simultaneously (e.g., rainy and hazy night), while they deal with each weather degradation individually ignoring the hybrid characterization. Therefore, to more pragmatically tackle real-world
adverse weather conditions, it is necessary and imminent to design a unified model that can handle multiple hybrid weather scenarios in an all-in-one fashion.

\begin{figure}
	\begin{center}
		\includegraphics[width=\linewidth]{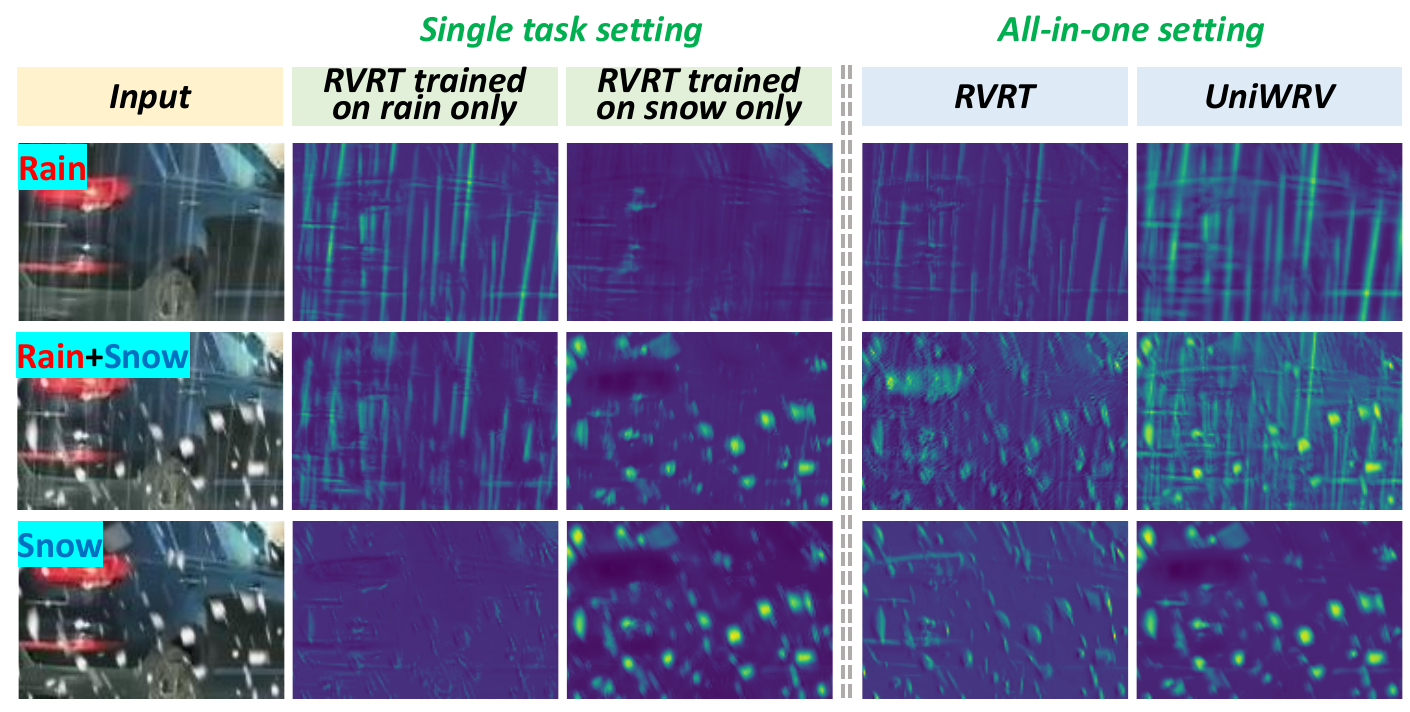}
	\end{center}
	\vspace{-5pt}
	\caption{Feature visualization of three heterogeneous weather conditions (i.e., rain, snow, and rain+snow). The conventional method RVRT \cite{liang2022recurrent} only perform well on single degradation learning however is unable to cope with unseen weather degradations. Directly applying RVRT to learn multiple weather degradations will lead to inadequate and ill-focused feature representation due to limited adaptation capacity. In contrast, our proposed UniWRV exhibits excellent adaptation capacity to heterogeneous degradations, and thus can adaptively and comprehensively handle given weather degradations.}
	\label{figure_hvis}
	\vspace{-10pt}
\end{figure}

The principal challenge towards this goal arises from the distribution heterogeneity of various uncertain video weather degradations during learning. The presence of various weather artifacts and their combinations introduces a broad spectrum of spatial degradation morphology and temporal degradation motion patterns across adjacent frames. Unified feature representation and modeling of multiple heterogeneous weather degradations necessitate the model to adaptively adapt to mine tailored degradation features upon input degradation characteristics rather than dealing with heterogeneous degradations with the same strategy. Existing video restoration methods can perform nicely in learning a single distribution of degradation, while becomes considerably challenging to delve into multiple heterogeneous degradations unified learning due to limited adaptability. This is mainly attributed to that the restricted adaptation capacity can lead to the model being optimized to a fixed compromise solution for heterogeneous degradations restoration which is sub-optimal for each individual degradation. In this case, the model fails to adapt for heterogeneous degradation to mine degradation-specific exclusive features, resulting in inadequate and ill-focused feature representation (Fig. \ref{figure_hvis}).
Consequently, to accomplish robust unified multiple hybrid adverse weather removal, an ideal unified model should possess the adaptation capability to discern and exploit the unique characteristics and properties inherent to different weather degradations and adaptively adjust its restoration tactics to perform tailored feature processing attuned to the given weather degradations.

To tackle the aforementioned problem, we propose a novel Unified model for multiple hybrid adverse Weather Removal from Video, termed UniWRV, which can automatically adapt to uncertain weather degradations and robustly generalize across various heterogeneous degradations (Fig. \ref{figure_hvis}). 
Specifically, we investigate the adaptation capability of the model to heterogeneous degradations in the spatial and temporal dimensions. To adaptively characterize heterogeneous spatial features of different weather videos, we propose a tailored weather prior guided module (WPGM) that can query exclusive weather priors for different instances as prompts to steer the feature characterization. Notably, the proposed WPGM module can be plug-and-play into existing image restoration algorithms to facilitate their implementation on all-in-one image restoration. 
In a similar vein, to adaptively fuse heterogeneous temporal features of different weather videos, we propose a dynamic routing aggregation module (DRA) that automatically selects optimal fusion paths for input video frames to perform dedicated multi-frame feature aggregation. Instead of depending on multi-path architecture with dear parameters and computations, DRA is equipped with a sparse modify weights selection scheme to achieve efficient dynamic fusion path routing with negligible additional computation and parameters. 

Another obstacle to reaching this goal is the absence of paired video data to simulate real-world hybrid adverse weather conditions. In this regard, we managed to create the first synthetic video dataset to facilitate multiple Hybrid adverse Weather removal from Video, dubbed HWVideo, which covers $15$ adverse weather conditions combined from three common weather factors (i.e., haze, rain, snow) and a crutial environmental factor (i.e., night). HWVideo includes 1200 training videos and 300 test videos, with a total of $900K$ adverse-weather/clean frame pairs.
Real-world adverse weather videos are also collected to further evaluate the generalizability and migration capability of the model.
Extensive experimental results substantiate the superiority of the proposed UniWRV for all-in-one multiple hybrid adverse weather removal from video. Furthermore, our approach can also be migrated to other all-in-one video restoration tasks with heterogeneous video degradations (i.e., unified video denoising, video deblurring, and video demoiréing) and outperform existing methods by a significant margin.

In conclusion, the main contributions are summarized as follows:
\begin{itemize}
	\setlength{\leftmargin}{-8pt}
	\item We propose a novel unified restoration model, namely UniWRV, to remove multiple hybrid adverse weather from video in an all-in-one fashion, which exhibits prominent adaptation capacities on heterogeneous video degradations learning.
	\item A weather prior guided module (WPGM) and a dynamic routing aggregation (DRA) module are proposed to handle adaptive spatial feature extraction and adaptive temporal feature fusion respectively. 
	\item A new synthetic video dataset HWVideo is constructed to more comprehensively simulate and characterize the complicated and unpredictable real-world hybrid weather scenarios.
	\item Our UniWRV achieves state-of-the-art performance on all-in-one multiple hybrid adverse weather removal from video as well as various other generic all-in-one video restoration tasks, outperforming existing methods by a significant margin. 
\end{itemize}
\vspace{-5pt}

The remaining of this paper is organized as follows: Section \ref{2} reviews existing Adverse Weather Removal methods and summarizes the popular prompt learning, prior learning, and dynamic neural networks. Section \ref{3} presents the methodology of how to design an adaptive unified video restoration model that can remove multiple hybrid adverse weather in an all-in-one fashion. Section \ref{4} demonstrates experiments to verify the performance of UniWRV on various application scenarios. Lastly, Section \ref{5} provides concluding remarks.

\section{Related Work}\label{2}
In this section, we start by providing a succinct overview of existing adverse weather removal literature. Then, we discuss the concept of prompt learning, prior learning, and dynamic neural networks that are tamed to implement model adaption capacity on heterogeneous degradation learning.
\subsection{Adverse Weather Removal.}
Adverse weather removal aims to restore crisp and clean backgrounds from corrupted images degraded by rain streaks \cite{wang2020model,fu2017removing,liu2018erase,wan2022image}, haze \cite{qin2020ffa,wu2021contrastive,zhang2021learning}, snow \cite{liu2018desnownet,chen2021all}, etc. The existing adverse weather removal algorithms can be roughly categorized as single-image-based \cite{zamir2021multi,zamir2022restormer,chen2021hinet,chen2022simple} and video-based ones \cite{ren2017video,wei2017should,wang2021seeing,zhang2021learning,yang2022learning,liu2018erase} depending on the input data.
In this work, we focus on removing complicated and diverse adverse weather from consecutive video frames, since this explores not only spatial features in a single image but also temporal correlations of adjacent frames.
Meanwhile, video is quite commonly used and easily accessible in real-world vision systems, and the inherent temporal information is desperately favored by many advanced downstream vision applications such as autonomous driving \cite{liang2018deep,prakash2021multi} and surveillance \cite{perera2018uav}. 
The early weather video restoration attempts focused on removing weather effects by designing specific models and networks for specific weather \cite{ren2017video,wei2017should,azizi2022salve,wang2021seeing,zhang2021learning,yang2022learning,liu2018erase,li2019video,liu2018d3r,yang2019frame,yue2021semi,yang2021recurrent,zhang2022enhanced}. However, since these methods are purposely designed for specific weather types, significant performance degradation can be observed when they are migrated to other tasks. Later, a series of investigations \cite{wang2019edvr,chan2021basicvsr,chan2022basicvsr++,liang2022recurrent,liang2022vrt,huang2022neural,li2023simple} intend to explore generic networks for restoring different degradations with identical architecture. Nevertheless, they still require independently trained weights for each task, hindering its generalizability in real-world applications.
Very recently, in the single image weather removal field, several unified models \cite{chen2022learning,li2020all,li2022all,valanarasu2022transweather,ozdenizci2023restoring,zhu2023learning,park2023all,zhang2023ingredient,yang2023visual,potlapalli2023promptir,ma2023prores,cheng2023deep,wan2023restoring} are proposed to remove multiple weather in an all-in-one fashion, which further improves the flexibility and deployability of the restoration model.
Though these methods work well for single-image weather removal, they will suffer from inferiority in processing video data due to the lack of temporal knowledge. Furthermore, real-world adverse weather conditions often suffer from simultaneous effects of multiple weather (e.g., rainy and hazy night),
while existing video restoration algorithms envisage that weather degradations occur independently, which may fail to handle real-world complicated scenarios. While the pioneering ViWS-Net \cite{yang2023video} addresses video multiple-type weather removal via weather messenger and adversarial learning, hybrid weather degradations are still unexplored. Despite BID \cite{han2022blind} and CPNet \cite{wang2023context} have made efforts in blind decomposition of mixed weather degradations, they fail to deal with non-decoupled weather like nighttime and only focus on a single image.
On the contrary, in this paper, we propose a convenient unified framework to remove stochastically occurring multiple hybrid adverse weather from video in an all-in-one fashion, thus being more desirable and feasible in real-world outdoor system applications. 

\begin{figure*}[t]
	\begin{center}
		\includegraphics[width=\linewidth]{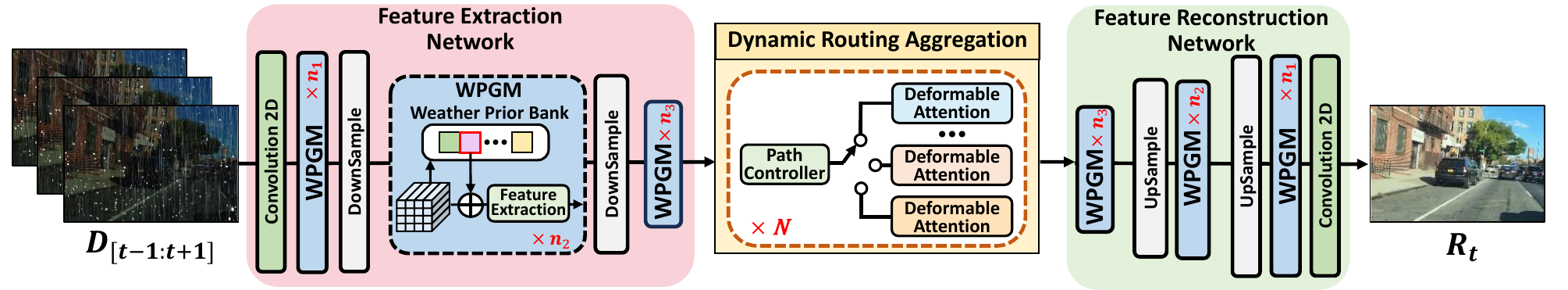}
	\end{center}
	\vspace{-7pt}
	\caption{ Architecture of UniWRV for unified multiple hybrid weather removal from video, which takes three adjacent frames with unknown weather artifacts as input and restores the clean mid-frame. Our UniWRV consists of three sub-modules: a feature extraction network and a feature reconstruction network built upon weather prior guided module (WPGM) that perform adaptive spatial features processing, a dynamic routing aggregation module (DRA) that performs adaptive multi-frame temporal features fusion. }
	\label{figure2}
	\vspace{-7pt}
\end{figure*}

\subsection{Prompt Learning and Prior Learning.}
Prompt engineering emerged from natural language processing community \cite{brown2020language,liu2023pre} aims to fine-tune pre-trained large language models to fit downstream tasks by training a small number of learnable parameters as prompts (i.e., a set of trainable vectors). These learned prompts can provide task-specific contextual knowledge to guide the model to perform more tailored feature operations \cite{zhou2022learning,li2021prefix}.
This highly desired property makes prompt an efficient and functional model "commander", and has been successfully applied in various non-single visual tasks like incremental learning \cite{wang2022dualprompt,wang2022learning} and multitask learning \cite{he2022hyperprompt,sohn2023visual}. 

Like-mindedly, prior is the inherent characteristic or experience of specific data summarized from learning or statistics. Downstream tasks can be facilitated by the pre-learned prior to easing the issue's complexity and difficulty \cite{gu2020image,pan2021exploiting,chen2021pre,li2021efficient}. Initially, the researchers in the image processing field \cite{he2010single,chen2013generalized,rudin1992nonlinear} mainly design exclusive prior manually based on competent observations of different data. However, these methods usually present inferior generalization and robustness due to competent assumptions and ill-posed parameters. Different
from these handcrafted prior, recent approaches seek to learn the task intrinsic prior knowledge implicitly through the network and optimize the prior continuously during training \cite{ulyanov2018deep,gu2020image,pan2021exploiting}. One category of these methods \cite{ulyanov2018deep,gu2020image,pan2021exploiting} recovers the corrupted portions by leveraging the image prior embedded in the pre-trained GAN. While another category of these methods \cite{chen2021pre,li2021efficient,liu2022tape} learns the statistical information of degraded images and clean images in advance by pre-training utilizes the learned prior to facilitate the modeling and characterization of the degradation. 

Intrigued by the desirable prior that implies specific degradation properties and desirable prompt that takes control of the model guidance, we advocate learning specific priors for different weather scenarios as prompts and querying the corresponding priors during inference to steer the model to adaptively perform dedicated spatial feature extraction and representation of heterogeneous degradations. One related work AWRCP \cite{ye2023adverse} proposes to reconstruct potentially clean images by utilizing high-quality image coding stored in the pre-trained codebook which is able to recover realistic texture details. However, AWRCP aims to utilize clean image features stored in the pre-trained codebook to fuel image restoration whereas our approach aims to provide guidance to the restoration network by storing prior information about different degradations in banks and thus facilitates adaptive perception of uncertain degradation which is more dedicate for coping with uncertain adverse weather videos in an all-in-one fashion.

\subsection{Dynamic Neural Networks.}
Dynamic neural networks \cite{wang2018skipnet,veit2018convolutional,jia2016dynamic,dai2017deformable,zhou2019spatio,gao2019deformable,su2016leaving,wu2019adaframe} are intended to adaptively adjust their weights or structure to handle given input with appropriate states, which often yield better compatibility, adaptiveness, and generalizability.
One category of
these studies \cite{wang2018skipnet,veit2018convolutional,jia2016dynamic,liu2018d3r,zhou2021trar,li2020learning} is devoted to dynamic routing for multi-branch or tree structures 
built on super networks, where each option path is a feasible data processing pipeline.
With dynamic routing for different examples, the networks can allow for more comprehensive and adequate characterization of diverse and differentiated data.
Inspired by these pioneer works, we pursue to dynamically capture the heterogeneous relationship between different weather adjacent frames by routing cross aggregation nodes, each of which performs its own role to fuse adept frame features. As will be
demonstrated, dynamic routing among fusion modules is reasonable and effective to tackle multi-frame feature fusion with uncertain heterogeneous weather degradations.

\section{Methodology}\label{3}
The overall framework of the proposed UniWRV is illustrated in Fig. \ref{figure2}. Our main goal is to develop an adaptive unified video restoration model that can remove multiple hybrid adverse weather in an all-in-one fashion. The core imperative to achieving this destination is how to make the network adaptively adapt to handle uncertain heterogeneous video features, comprehensively mining tailored degradation features of given weather conditions and thus approximate potentially optimal restoration tactics for each individual degradation. 
To this end, we explore two dimensions of adaptivity to tackle the heterogeneity of uncertain video features: spatial adaptivity and temporal adaptivity. 
To adaptively handle heterogeneous spatial features of uncertain weather videos, we propose a weather prior guided module (WPGM) that queries exclusive priors for each given instance as prompts to steer the subsequent feature extraction operation, thus allowing more tailored representation for uncertain weather conditions. Meanwhile, to adaptively handle heterogeneous temporal features of uncertain weather videos, we propose a dynamic routing aggregation module (DRA) to select the optimal temporal feature fusion path for each given instance to flexibly integrate multi-frame features of different weather conditions. 

As shown in Fig. \ref{figure2}, our UniWRV contains three components: the feature extraction network, the dynamic routing aggregation module, and the reconstruction network, where the feature extraction and reconstruction networks are built upon WPGM with pixel-unshuffle and pixel-shuffle operations for downsampling and upsampling purpose. 
Formally, given three
consecutive degraded frames $D_{[t-1:t+1 ]}$, the feature extraction network first produces the deep extracted features $F_{[t-1:t+1 ]}$. Thereafter, DRA adaptively aggregates $F_{[t-1:t+1]}$ via dynamic routing and generates the fused middle frame feature $M_N$. 
Finally, $M_N$ are input into the feature reconstruction network to restore the clean middle frame $R_t$ for $D_t$.

\subsection{Weather Prior Guided Module}

The spatial characteristics inherent to different weather degradations exhibit substantial heterogeneity in terms of morphology and appearance. Simply learning generic representations of different degradations via conventional static feature processing operations fails to capture unique and exclusive spatial features of specific weather degradations due to the lack of adaption capacity. Therefore, it is crucial for the unified to discern the degradation properties and adaptively adjust to characterize the heterogeneous features of given weather degradations.
To address the challenge of accommodating spatial heterogeneity intrinsic to uncertain weather degradations, we propose to identify the weather degradation properties of a given instance via weather prior query and utilize the queried priors as prompts to steer the spatial feature characterization. In practice, weather priors that represent different weather degradation properties are learned and stored in a weather prior bank, each learned weather prior vector summarizes a class of feature distributions with similar weather degradation properties. The weather prior and the subsequent feature extraction module are optimized together during the training process, the feature extraction module is responsible for learning the prompts embedded in the prior and incorporating them into the extraction process of the input video feature. The added prior controls the restoration of degraded video by establishing complex interactions with video features through the feature extraction module which not only avoids the introduction of additional modules but also allows the network to build complex interactions adaptively. During inference, the most matching prior is queried as a prompt by measuring the distance between the pending feature and the weather priors in the bank (smaller distances indicate more similar degradation properties), and subsequent feature extraction operations can utilize this prompt to identify the spatial degradation properties of the pending features and adaptively perform corresponding feature characterization.

\begin{figure*}[t]
	\begin{center}
		\includegraphics[width=\linewidth]{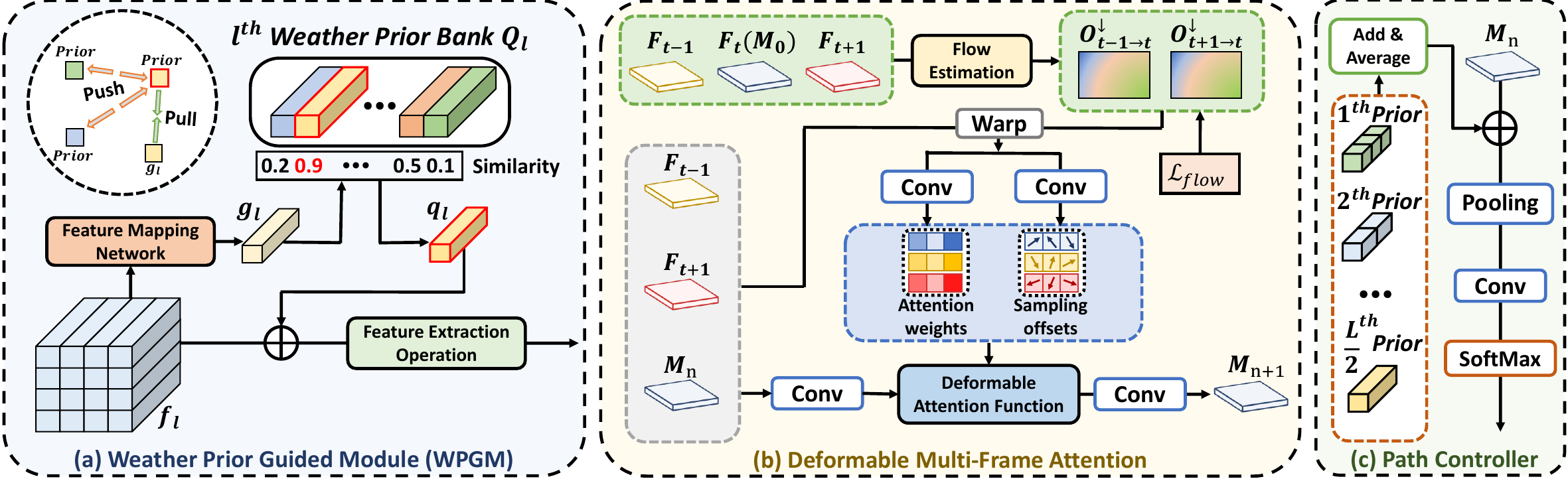}
	\end{center}
	\vspace{-7pt}
	\caption{ (a): Weather prior guided module that performs weather adaptive feature extraction via prior query. (b): Deformable multi-frame attention block that performs multi-frame feature fusion. (c): Path controller that determines the most appropriate fusion routing of multi-frame features. }
	\label{figure2_1}
	\vspace{-5pt}
\end{figure*}

The schematic illustration of WPGM is depicted in Fig. \ref{figure2_1}(a). Given an input feature $f_l \in \mathbb{R}^{H \times W \times C}$ of $l^{th}$ layer, the feature mapping network first embeds it as a global compressed vector $g_l \in \mathbb{R}^{C}$ in latent space. Two simple fully connected layers with ReLU activation function are applied to construct the mapping network. Then, a quantization $q(\cdot)$ is performed on the latent vector $g_l$ to locate the most similar weather prior vector $q_l \in \mathbb{R}^{C}$ in the weather prior bank $Q_l=\{q_l^i\}_{i=1}^{P_n}$ with $P_n$ entries (20 by default):
\begin{equation}
	\begin{aligned} 
		q_l = q(g_l) := arg min_{q_l^i \in Q_l} \parallel g_l - q_l^i \parallel,
	\end{aligned}
\end{equation}
where $q(\cdot)$ indicate nearest neighbor matching operation. Considering that samples of the same weather type may have different degradation characteristics (e.g., dense rain or snow exhibits the characteristics of haze), which is also consistent with the one discussed in Fig. \ref{figure_hvis}, the heterogeneous weather degradation contains both exclusive degradation components as well as sharing certain degradation components. Thus, we do not explicitly constrain the weather prior to be one-to-one with the weather type, but instead implicitly make them correspond to features with similar degradation properties through the distance measure. Such a strategy allows for more comprehensive and accurate modeling of diverse heterogeneous weather scenarios. Furthermore, instead of combining all the vectors by soft weights to obtain continuous prior vector, which will increase the difficulty of prior modeling and make the network more difficult to decode the knowledge embedded in the queried weather prior \cite{razavi2019generating,esser2021taming}, we only select the discrete weather vector with the highest similarity in the prior bank. The queried weather prior $q_l$ is then broadcast added with input feature $f_l$ to propagate through the feature extraction operation. 
Notably, the feature extraction operation can be substituted with any existing mature blocks, and in this work, we deploy the simple residual block \cite{he2016deep} as a baseline model.

\noindent\textbf{Training objectives.} To train the feature mapping networks and weather prior banks, we first adopt a vector-level loss to reduce the distance between the embedded latent vectors $g_{[1:L]}$ and the queried weather priors $q_{[1:L]}$:
\begin{equation}
	\resizebox{.95\hsize}{!}{$
		\begin{aligned} 
			\mathcal{L}_{prior}^v =  \sum\nolimits_{l=1}^{L}[ 1-cos(q_l , sg(g_l))] + \beta \sum\nolimits_{l=1}^{L}[ 1-cos(sg(q_l) , g_l)],
		\end{aligned}$}
\end{equation}
where $L$ represents the total layers of WPGD and $sg(\cdot)$ stands for the stop-gradient operator. $\beta$ is set to 0.25 which performs update rates trade-off between mapping networks and weather prior banks. Apart from vector-level loss, we also employ a contrastive loss to maximize the consistency of the embedded latent vectors $g_{[1:L]}$ and the queried weather priors $q_{[1:L]}$ (i.e., positive samples),
while minimizing the consistency between $g_{[1:L]}$ and other weather priors $q_{[1:L]}^{[1:P_{n}-1]-}$ in the weather prior banks (i.e., negative samples). The contrastive loss is designed to create distinct regions within the representation space and facilitate a clearer separation between individual weather priors, ensuring that different weather conditions can be recognized accurately rather than being misclassified due to small deviations. The contrastive loss can be formulated as:
\begin{equation}
	\resizebox{.75\hsize}{!}{$
		\begin{aligned} 
			\mathcal{L}_{prior}^c = \sum\nolimits_{l=1}^{L}[-log\frac{exp(g_l \cdot q_l / \tau )}{\sum\nolimits_{i=1}^{N-1}exp(g_l \cdot q^{i-}_l / \tau)}],
		\end{aligned}$}
\end{equation}
where $\tau$ is a temperature hyper-parameter set to 0.07 by default.

\begin{figure}
	\centering
	\includegraphics[width=.9\linewidth]{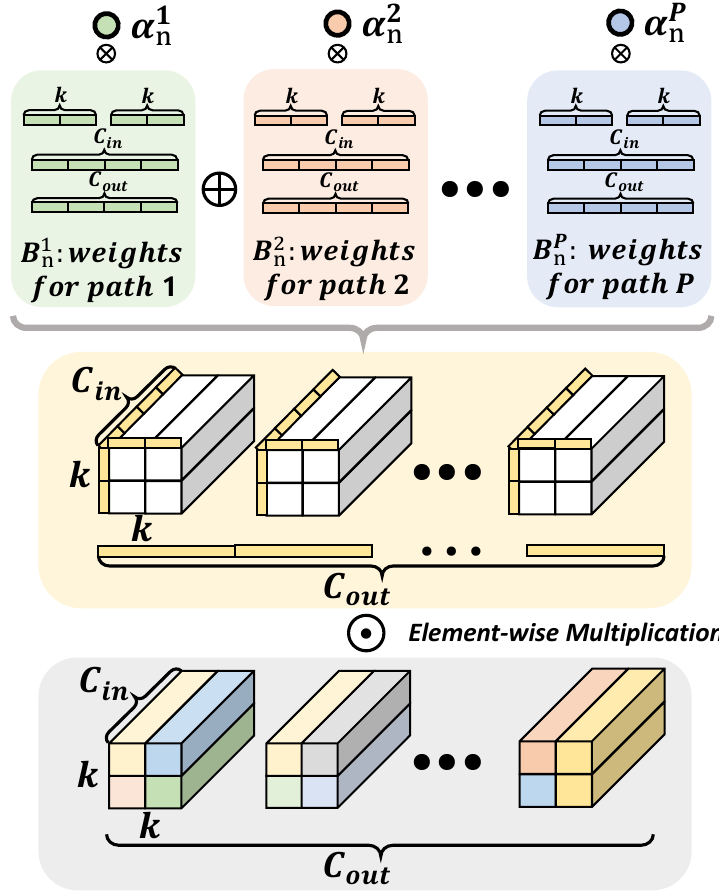}
	\vspace{-4pt}
	\caption{Illustration of the proposed modify weights routing scheme. Instead of routing via redundant multi-node structure, we advocate routing cross lightweight modify weights and performing only one projection calculation with the modified parameter.}
	\label{figure3}
	\vspace{-11pt}
\end{figure}

\subsection{Dynamic Routing Aggregation}
Apart from the spatial domain, another core dimension of video data is temporal correlation. In the temporal domain, different weather videos enjoy heterogeneous characteristics in the movement pattern within adjacent frames, the manners of valuable information mining from adjacent frames can be distinctive in different weather conditions (e.g., the model can extract features of the obscured area from adjacent frames for rainy videos while focusing more on illumination changes for nighttime videos). Consequently, the adjacent frame feature fusion mechanism of different weather videos is heterogeneous, and the model should select customized fusion strategies for distinct degradation videos to mine degradation-specific adjacent frame relationships. To adaptively fuse adjacent frames of uncertain weather videos, we propose to dynamically select optimal fusion strategies for different instances through a dynamic routing scheme, which can flexibly fuse adjacent frame features by routing across learned fusion nodes. 

To achieve dynamic routing, an intuitive solution is to construct a multi-branch structure, where each branch is equipped with different projection weights for processing different weather videos. However, such a design will inevitability introduce additional unexpected parameters, which would dramatically exacerbate the computational overhead and inhibit its deployment in real-world vision systems. To ameliorate this limitation, our main goal is to design a routing scheme that can select exclusive paths for each instance while being lighter-weight on parameter size and computational burden. Subsequently, we will progressively analyze how to satisfy this demand. Specifically, for the n-th routing layer, given the input features $M_n$ and routing space $S_n^{[1:P]}$ (path nodes in the n-th layer), where $P$ denotes the number of optional paths, the output features of a vanilla routing solution can be obtained by:
\begin{equation}
	\begin{aligned} 
		M_{n+1} = \sum\nolimits_{i=1}^{P} \alpha_n^i S_n^i(M_n,W_n^i),
	\end{aligned}
	\label{eq3}
\end{equation}
where $\alpha_n^{[1:P]}$ are the path probabilities predicted by the
path controller and $W_n^{[1:P]}$ represent the filters parameters of path nodes. In this work, we employ continuous $\alpha$ to obtain a soft routing, which can also be binarized for a hard
path routing via Gumbel softmax \cite{jang2016categorical} but with inferior performance (see Sec. \ref{abla}). It can be easily observed from Eq. \ref{eq3} that the computational burden of the vanilla routing scheme mainly originates from the multi-node computation, i.e., each node has to perform projection on $M_n$ individually. This triggers us to explore whether the multi-node computation can be implemented with only one operation. By revisiting the working mechanism of multi-node routing, we found that routing can operate on the node parameters $W_n^{[1:P]}$ and then perform a single projection calculation with the routed parameter $W_n$, which can be formulated as:
\begin{equation}
	\begin{aligned} 
		M_{n+1} =   S_n(M_n,\sum\nolimits_{i=1}^{P}\alpha_n^i W_n^i).
	\end{aligned}
	\label{eq4}
\end{equation}
However, multiple independent parameters $W_n^{[1:P]}$ still need to be stored and learned which also delivers an undesired memory footprint. Therefore, to alleviate the parameter burden, we further simplify the parameter routing to a sparse modify weights routing, i.e., routing among sparse modify weights $B_n^{[1:P]}$ rather than the full-size parameters $W_n^{[1:P]}$. More concretely, each modify weight $B_n^{i}$ only contains four vectors with lengths equal to the four dimensions of convolution kernel parameters ($k$, $k$, $C_{in}$, and $C_{out}$). We perform dynamic routing on the sparse modify weights and multiply the routed weight with the single base parameter $W_n$ to obtain the modified parameter.
The overall process is illustrated in Fig. \ref{figure3} and the n-th routing layer is then defined as:
\begin{equation}
	\begin{aligned} 
		M_{n+1} &=   S_n(M_n,\sum\nolimits_{i=1}^{P}\alpha_n^i B_i W_n),\\ B_{i} &= \{k^i,k^i,C_{in}^i, C_{out}^i \}.
	\end{aligned}
	\label{eq5}
\end{equation}
This is a macroscopic mathematical representation of the n-th routing layer and the detailed operations of $S_n$ will be introduced in the next section.
The finally proposed routing scheme significantly reduces the computational complexity and parameter size, and the goal of dynamic routing to adaptively aggregate uncertain multi-weather frame features can still be accomplished.

\noindent\textbf{Deformable multi-frame attention as path node.} Considering that different frames are affected by different areas and degrees of weather, we propose a deformable multi-frame attention to adaptively generate sampling points for neighboring frames as well as the weights of the sampling points to efficiently fuse the valuable features of neighboring frames. Besides, the unaligned neighboring frames are warped by optical flow prediction to more accurately predict the expected attention weights and sampling offsets, delivering more effective multi-frame feature mining and fusion. As shown in Fig. \ref{figure2_1}(b), our proposed DRA aggregates desirable features from adjacent frames through deformable multi-frame attention blocks (i.e., the routing node), which successively optimize fusion features $M_n$ through mining desirable features from adjacent frame features $F_{t-1}$ and $F_{t+1}$ over $N$ routing layers. The above designed sparse parameter routing scheme is deployed to each convolution operation in the block to achieve dynamic routing. Given the features $F_{[t-1:t+1]} \in \mathbb{R}^{\frac{H}{4} \times \frac{W}{4} \times 4C}$ of three consecutive frames, the flow estimation network first predicts the optical flow $O_{t-1\rightarrow t}^{\downarrow}$ and $O_{t+1\rightarrow t}^{\downarrow}$, which are utilized to warp the adjacent frame features to the central one:
\begin{equation}
		\label{q7}
	\begin{aligned} 
		F_{t-1}^w &= Warp(F_{t-1},O_{t-1\rightarrow t}^{\downarrow}),\\ F_{t+1}^w &= Warp(F_{t+1},O_{t+1\rightarrow t}^{\downarrow}).
	\end{aligned}
\end{equation}
Convolution operations are deployed to project the warped features and obtain the attention weights $A_n \in \mathbb{R}^{\frac{H}{4}\times \frac{W}{4} \times M  T K}$, sampling offsets
$\Delta P_n \in \mathbb{R}^{\frac{H}{4} \times\frac{W}{4} \times M  T K}$, and projected features $M_n^f \in \mathbb{R}^{\frac{H}{4} \times \frac{W}{4}\times 4TC}$, where $M$, $T$, and $K$ represent the number of attention heads, the number of frames in the sliding window, and the number of sampling points, respectively.
\begin{equation}
		\label{q8}
	\begin{aligned} 
		A_n &= Conv(Cat(F_{t-1}^w, M_n,F_{t+1}^w)),\\
		\Delta P_n &= Conv(Cat(F_{t-1}^w, M_n,F_{t+1}^w)),\\
		M_n^f &= Conv(M_n,F_{t-1},F_{t+1}).
	\end{aligned}
\end{equation}
$A_n$, $\Delta P_n$, and $M_n^f$ are then fed to the deformable attention function
$DefAtt(\cdot)$ \cite{zhu2020deformable} and produces the aggregated output features $M_{n+1}$ of the n-th routing layer via a convolution layer:
\begin{equation}
	\label{q9}
	\begin{aligned} 
		M_{n+1} = Conv(DefAtt(A_n,\Delta P_n,M_n^f)).
	\end{aligned}
\end{equation}
Eq. \ref{q7}, Eq. \ref{q8}, and Eq. \ref{q9} inscribe the specific operational operators of $S_n$ in Eq. \ref{eq5} and all their parameters (i.e., convolution kernels) are dynamiclly routed via the defination in Eq. \ref{eq5}.

\noindent\textbf{Path controller.} In DRA, each frame feature fusion layer is equipped with a path controller (Fig. \ref{figure2_1}(c)) to determine the routing selection for given examples. Specifically, the path controller first adds and averages the weather prior vectors queried by different layers in the feature extraction network to an aggregated prior vector. The obtained aggregated prior and the input feature $M_n$ are then broadcast summed and transmitted to a global pooling layer and a convolutional layer. Finally, the generated features are mapped by softmax to derive the routing weights $\alpha_n$.
Mathematically, such a process can be defined as:
\begin{equation}
	\begin{aligned} 
		\alpha &= Softmax(Conv(Pool(M_n^p))),\\
		M_n^p &=	M_n + \frac{rep(1^{th} Prior) + \cdots + \frac{L}{2}^{th} Prior }{\frac{L}{2}},
	\end{aligned}
\end{equation}
where $rep(\cdot)$ denotes repeating the low-dimensional vectors ($1^{th} Prior:(\frac{L}{2}-1)^{th} Prior$) to the same length as the high-latitude vector ($\frac{L}{2}^{th} Prior$).

\noindent\textbf{Training objective.} To train the flow estimator, a warp loss is introduced, which is computed as:
\begin{equation}
	\begin{aligned} 
		\mathcal{L}_{flow} = &\parallel G_t^{\downarrow} - Warp(G_{t-1}^{\downarrow}, O_{t-1\rightarrow t}^{\downarrow} ) \parallel^2 + \\
		&\parallel G_t^{\downarrow} - Warp(G_{t+1}^{\downarrow}, O_{t+1\rightarrow t}^{\downarrow} ) \parallel^2,
	\end{aligned}
\end{equation}
where $G_{t-1}^{\downarrow}$, $\parallel G_t^{\downarrow}$, and $G_{t+1}^{\downarrow}$ are the three downsampled frames and ``Warp” denotes the warp operation.

\subsection{Total training objectives.} After presenting the proposed two core designs, we give the complete loss below. Apart from the aforementioned losses: the weather prior losses $\mathcal{L}_{prior}^v$ and $\mathcal{L}_{prior}^c$, flow warp loss $\mathcal{L}_{flow}$, a $\mathcal{L}_1$ loss is also employed to minimize the distance between the restored mid-frame $R_t$ and ground truth $G_t$, and the complete objective of our UniWRV is defined as: 
\begin{equation}
	\begin{aligned} 
		\mathcal{L} = \mathcal{L}_1(R_t,G_t) + \mathcal{L}_{prior}^v  + \mathcal{L}_{prior}^c + \mathcal{L}_{flow}.
	\end{aligned}
\end{equation}

\subsection{Hybrid Weather Video Dataset}

In order to train the proposed UniWRV network, we managed to create a synthetic hybrid adverse weather video dataset (HWvideo), which contains 15 scenarios derived from random combinations of three common weather factors (i.e., haze, rain, snow) and a crutial environmental factor (i.e., night). 
For the fabrication, we first manually curate 100 well-illuminated
video of normal weather scenarios from BDD100K \cite{yu2018bdd100k} autonomous driving database as ground truth, where each video is randomly clipped to a 20-second piece. Since the BDD100K contains the interior scenes of the car at the periphery of the video at the time of filming, we further cropped out the central $720\times480$ sub-video of each piece to avoid such effort which includes only the natural sceneries outside the car. Then, we utilize the
commercial editing software Adobe After Effects \cite{ae} to create realistic synthetic weather effects for each video which were performed by five recruited post-processing experts. For each video, we synthesize the aforementioned 15 weather conditions with lighting conditions, scenes, subjects, and styles, e.g., taken into
account to ensure realism and diversity. 80\% of the synthetic data is divided into training sets and the rest is used for testing.
We also collected 50 real-world hybrid adverse weather videos from the Youtube website\footnote{https://www.youtube.com/} to evaluate the model performance under practical conditions as well as the generalization ability.

\begin{table}
	\vspace{-10pt}
	\centering
	\caption{Comparison of HWVideo against conventional video adverse weather removal datasets.}
	\tabcolsep=3pt
	\footnotesize
	\begin{tabular}{l|cccccc}
		\toprule
		Dataset& Haze & Rain& Snow & Night &Videos& Frames\\
		
		\midrule
		\makecell[l]{REVIDE  \cite{zhang2021learning} } &$\checkmark$ & &&&48&$\sim\!2K$\\
		\makecell[l]{NTURain \cite{chen2018robust} } & &$\checkmark$ &&&33&$\sim\!4.8K$\\
		\makecell[l]{RainSynLight25 \cite{liu2018erase} } & &$\checkmark$&&&215&$\sim\!2.5K$\\
		\makecell[l]{RainSynComplex25 \cite{liu2018erase} } & &$\checkmark$&&&215&$\sim\!2.5K$\\
		
		\makecell[l]{RainSynAll100 \cite{yang2021recurrent} } &$\checkmark$ &$\checkmark$ &&&1000&$8K$\\
		\makecell[l]{SMOID  \cite{jiang2019learning} } && &&$\checkmark$ &179&$35.8K$\\
		\makecell[l]{SDSD  \cite{wang2021seeing} } && &&$\checkmark$ &150&$37.5K$\\
		\makecell[l]{\textbf{HWVideo} } & $\checkmark$& $\checkmark$& $\checkmark$& $\checkmark$&\bm{$1500$} &\bm{$900K$}\\
		
		\bottomrule
	\end{tabular}
	\label{table_dataset}
	\vspace{-10pt}
\end{table} 

In Tab. \ref{table_dataset}, we compare HWVideo to other adverse weather video datasets. These datasets only focus on single or double degradations, while we consider four weather factors in combination to form 15 hybrid weather scenarios, which can more comprehensively simulate complicated real-world adverse weather conditions where multiple weather can coexist. This richness in diversity and volume not only facilitates a more robust evaluation of models but also enables reliable and practical applications in addressing real-world adverse weather conditions, especially in autonomous driving systems. In addition, the simple and efficient manner of obtaining the dataset allows it to be further extended in new scenarios (e.g., surveillance video). 
\vspace{-6pt}
\section{Experiments and Analysis}\label{4}

\subsection{Implementation Details}

All experiments are implemented with the Pytorch framework on four NVIDIA RTX 3090ti GPUs. We train our UniWRV for $6\times 10^5$
iterations using Adam as the optimizer with a batch size of 16.
The initial learning rate is $2 \times 10^{-4}$, which is steadily decreased to $1 \times 10^{-6}$ using the cosine annealing strategy \cite{loshchilov2016sgdr}. The training frames are randomly cropped to $256\times256$ and random flipping or rotation is applied for data augmentation. 
PSNR \cite{huynh2008scope} and SSIM
\cite{wang2004image} are utilized to evaluate the restoration performance. 
For network implementation, the number of channels for the feature extraction network and the feature reconstruction network at the three scales are 32, 64, and 128, respectively. The number of feature channels in dynamic routing aggregation is consistently set to 256. The length of the vectors in weather prior banks is the same as the channel of the corresponding scale feature maps. To achieve a better trade-off between video weather removal quality and computational efficiency, the M, K, and T are set as 4, 12, and 3, respectively. 
\subsection{Comparison with the State-of-the-Arts on HWVideo Dataset}

\begin{table*}[h]
	\setlength{\tabcolsep}{10pt}	
	\caption{Quantitative comparison with the SOTA methods on our proposed benchmark HWVideo dataset.
		Top super row: all-in-one single image weather removal methods trained and tested on all conditions with a single model.
		Middle row: video weather removal methods trained and tested separately on each weather condition. Bottom super
		row: video weather removal methods trained and tested on all conditions in an all-in-one fashion. Multiplicative numeral indicates the number of degradation types contained in a video (e.g., Triple represents the average scores of $C_4^3=4$ conditions containing three degradation types).}
	\begin{center}
		\begin{tabular}{c|cccccccc|cc}
			\toprule[1pt]
			\makecell[l]{\multirow{2}{*}{Method}} &\multicolumn{2}{c}{Single}&\multicolumn{2}{c}{Double}&\multicolumn{2}{c}{Triple}&\multicolumn{2}{c}{Quadruple}&\multicolumn{2}{|c}{Average}
			\\
			& PSNR & SSIM& PSNR & SSIM&PSNR & SSIM&PSNR & SSIM&PSNR & SSIM\\
			\midrule
			
			\makecell[l]{TransWeather \cite{valanarasu2022transweather}} &21.55&0.8019&18.59&0.7528&15.98&0.6498&14.29&0.6087&17.60& 0.7033 \\
			\makecell[l]{WeatherDiff \cite{ozdenizci2023restoring}} &21.87&0.8078&18.81&0.7577&15.94&0.6501&14.28&0.6078&17.73&0.7059\\
			\makecell[l]{AirNet \cite{li2022all}}  &22.12&0.8125&18.87&0.7585&16.02&0.6517&14.33&0.6101&17.84& 0.7082\\
			\makecell[l]{TKL \cite{chen2022learning}}  &22.16&0.8112&18.98&0.7599&16.18&0.6532&14.45&0.6117&17.94&0.7090 \\
			\makecell[l]{PromptIR \cite{potlapalli2023promptir}}  &22.36&0.8223&19.35&0.7635&16.45&0.6593&14.75&0.6203&18.23& 0.7164\\
			
			\midrule
			\makecell[l]{EDVR \cite{wang2019edvr}} &27.24&0.8599&24.33&0.8009&20.17&0.7691&17.53&0.6494&22.32&0.7698\\
			\makecell[l]{BasicVSR \cite{chan2021basicvsr}} &27.91&0.8633&24.47&0.8145&20.76&0.7784&17.78&0.6571&22.73&0.7783\\
			
			\makecell[l]{BasicVSR++ \cite{chan2022basicvsr++}} &28.28&0.8875&24.96&0.8251&21.25&0.7865&17.97&0.6625&23.12&0.7904\\
			\makecell[l]{VRT \cite{liang2022vrt}} &28.54&0.8915&25.23&0.8292&21.85&0.7925&18.44&0.6801&23.52&0.7983\\
			\makecell[l]{Shift-Net \cite{li2023simple}} &28.86&0.9002&25.73&0.8411&22.17&0.7965&18.62&0.6896&23.85&0.8069\\
			\makecell[l]{RVRT \cite{liang2022recurrent}} &28.98&0.9018&25.96&0.8469&22.27&0.8005&18.85&0.6933&24.02&0.8106\\
			\makecell[l]{ViWS-Net \cite{yang2023video}} &29.03&0.9059&26.28&0.8599&22.67&0.8165&18.99&0.7103&24.24&0.8232\\
			\makecell[l]{\textbf{UniWRV}} &\textbf{29.20} &\textbf{0.9132}&\textbf{26.83}&\textbf{0.8899}&\textbf{23.67}&\textbf{0.8497}&\textbf{20.80}&\textbf{0.7509}&\textbf{25.13}&\textbf{0.8509}\\
			\midrule
			\makecell[l]{EDVR \cite{wang2019edvr}} &23.19&0.8199&20.33&0.7985&17.67&0.7285&15.63&0.6254&19.21&0.7431\\
			\makecell[l]{BasicVSR \cite{chan2021basicvsr}} &23.86&0.8233&20.97&0.8085&18.16&0.7384&16.08&0.6271&19.77&0.7493\\
			
			\makecell[l]{BasicVSR++ \cite{chan2022basicvsr++}} &24.75&0.8296&21.57&0.8129&18.55&0.7435&16.27&0.6339&20.29&0.7550\\
			\makecell[l]{VRT \cite{liang2022vrt}} &25.18&0.8385&22.08&0.8122&19.17&0.7445&16.74&0.6402&20.79&0.7589\\
			\makecell[l]{Shift-Net \cite{li2023simple}} &25.48&0.8399&22.13&0.8133&19.21&0.7465&16.75&0.6396&20.89&0.7598\\
			\makecell[l]{RVRT \cite{liang2022recurrent}} &25.85&0.8587&22.16&0.8147&19.26&0.7582&16.97&0.6425&21.06&0.7685\\
			\makecell[l]{ViWS-Net \cite{yang2023video}} &26.45&0.8682&22.86&0.8248&19.61&0.7952&17.53&0.6765&21.61&0.7912\\
			\makecell[l]{\textbf{UniWRV}} &\textbf{28.48} &\textbf{0.8981}&\textbf{25.63}&\textbf{0.8546}&\textbf{22.37}&\textbf{0.8152}&\textbf{20.99}&\textbf{0.7866}&\textbf{24.37}&\textbf{0.8386}\\
			\bottomrule[1pt]
		\end{tabular} 
	\end{center}
	\label{table1}
	\vspace{-8pt}
\end{table*}
The quantitative results on the HWVideo dataset are reported in Tab. \ref{table1}. All the compared methods were trained and tested on the HWVideo dataset according to the same criteria for a fair and unbiased comparison. We trained the existing methods according to the code publicly available in their paper, and all of them were trained to stable convergence (i.e., the PSNR does not change more than 0.005 dB in three consecutive epochs). The table provides results for three types of experiments: (1) all-in-one single image weather removal methods trained and tested on all conditions with a single model; (2) video weather removal methods trained and tested separately on each weather condition; (3) video weather removal methods trained and tested on all conditions in an all-in-one fashion. As can be found, our UniWRV delivers unparalleled 
performance gains and outperforms all competitive models both in the condition-specific setting and in the all-in-one setting. While existing video restoration methods can perform reasonably well in task-specific scenarios (middle row), they exhibit stretched adaptability when dealing with multiple heterogeneous degradations learning (bottom row).
Notably, UniWRV exceeds the top-performing video restoration approach RVRT \cite{liang2022recurrent} by a significant margin of 3.31dB PSNR on average under the all-in-one setting. Furthermore, our method can robustly cope with highly ambiguous and complex hybrid weather scenarios, and the more weather degradation is combined in the video, the more obvious the advantages of our method become. Especially for ``Quadruple" conditions, our UniWRV surpasses the previous best ViWS-Net by a significant 3.46dB in PSNR.
For single-image weather removal algorithms, despite being specifically designed for handling multiple weather degradations \cite{valanarasu2022transweather,li2022all,chen2022learning}, they are still inferior to existing video restoration methods when processing video data. 

\begin{figure*}[t]
	\begin{center}
		\includegraphics[width=\linewidth]{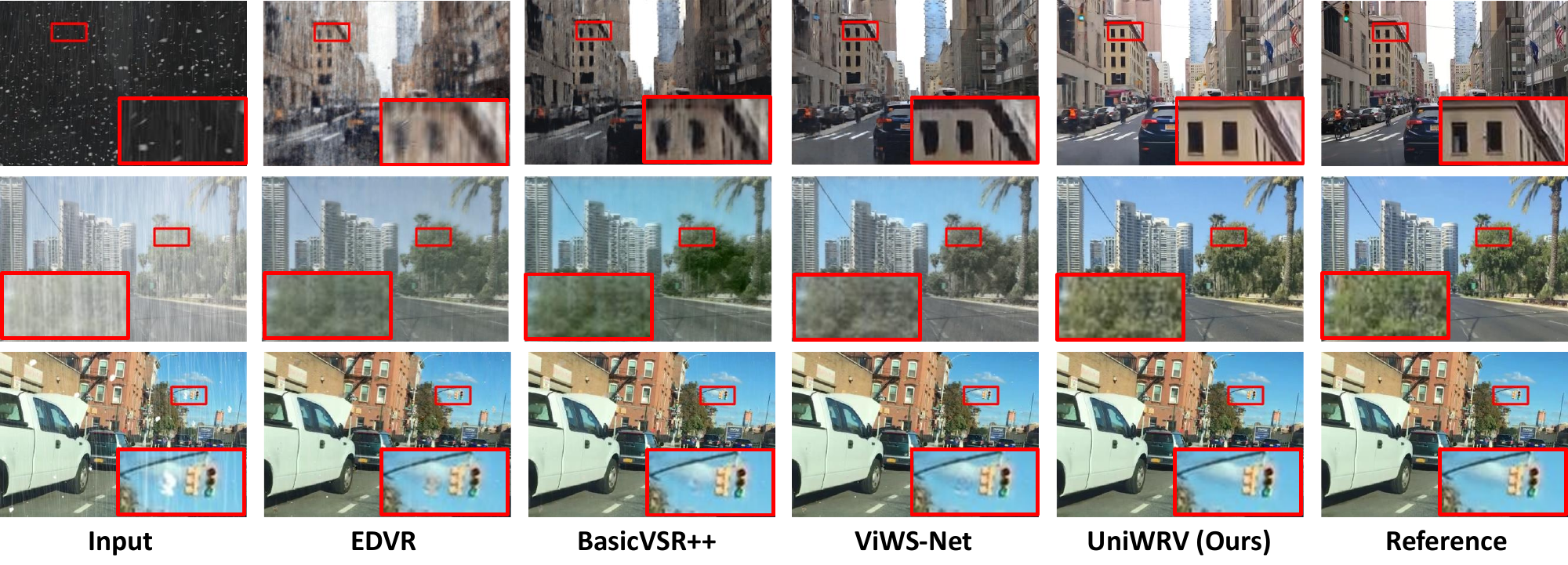}
	\end{center}
	\vspace{-7pt}
	\caption{ Visual comparisons with SOTA adverse weather removal methods on the HWVideo dataset. Reconstructed results demonstrate the ability of UniWRV to restore arbitrary hybrid adverse conditions while preserving more details, hence yielding more visually pleasing results.} 
	\label{figure4}
	\vspace{-5pt}
\end{figure*}

We also demonstrate visual comparisons in
Fig. \ref{figure4}. As suggested, UniWRV recovers more coherent fine structures with natural textures and consistent illumination while other methods tend to introduce inconspicuous details and unnatural colors, especially the scenario of four weather mixed together (top row).

\begin{figure*}[t]
	\begin{center}
		\includegraphics[width=\linewidth]{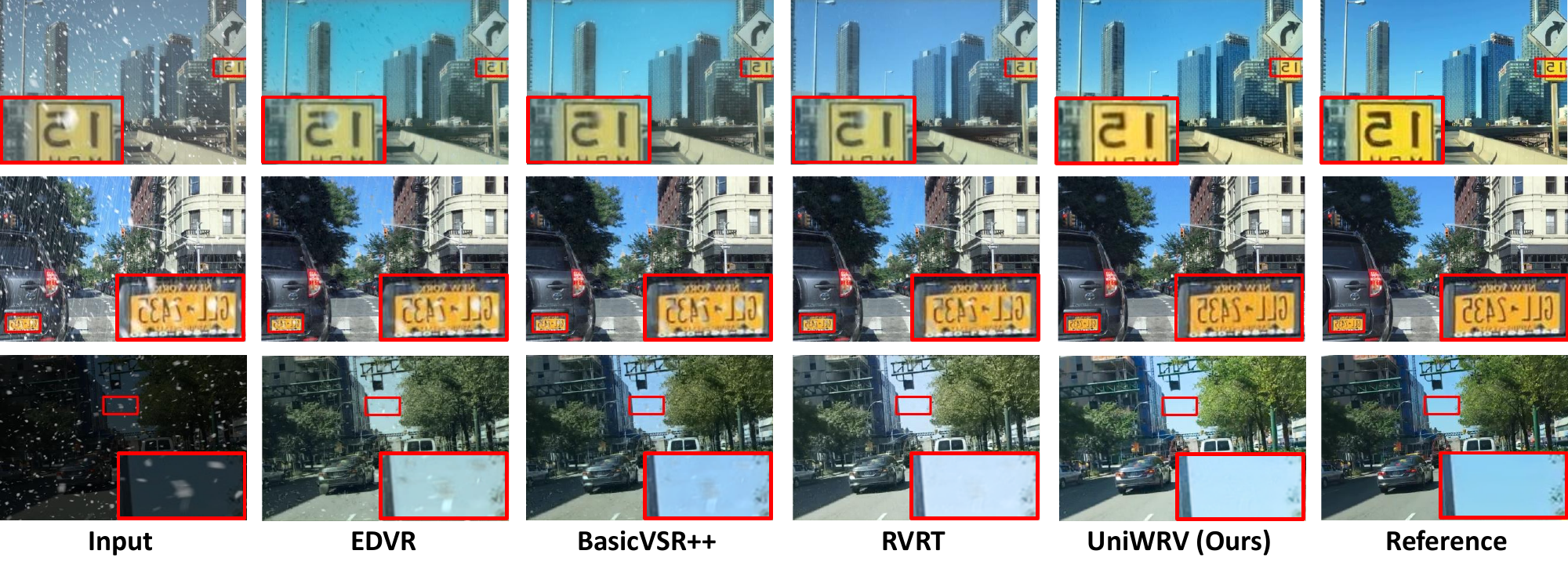}
	\end{center}
	\vspace{-7pt}
	\caption{ Visual comparisons with SOTA adverse weather removal methods on the HWVideo dataset.  }
	\label{figure4_1}
	\vspace{-10pt}
\end{figure*}

\subsection{Results on Real-world Conditions and High-level Applications}

\begin{figure}[t]
	\centering
	\includegraphics[width=\linewidth]{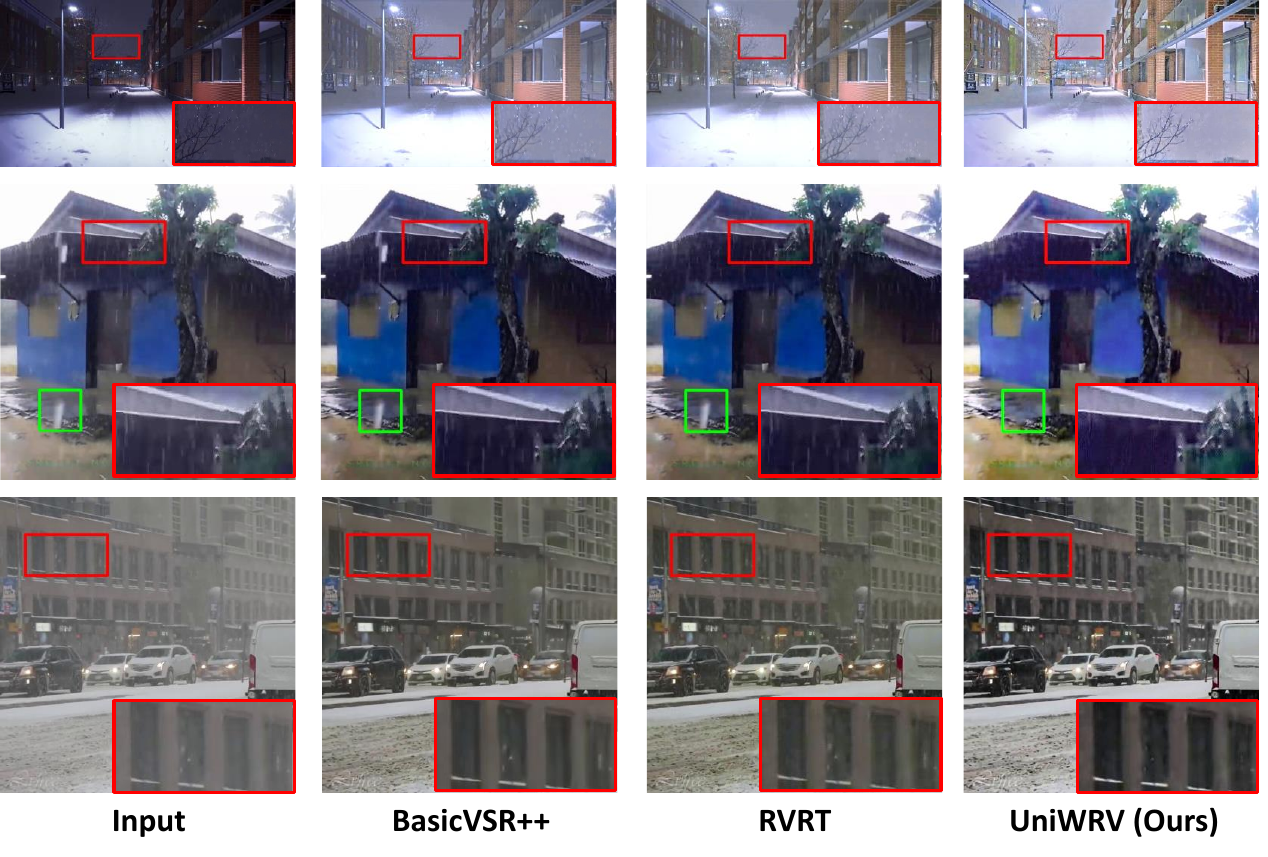}
	\vspace{-17pt}
	\caption{Four examples of real-world hybrid adverse weather removal results delivered by BasicVSR++ \cite{chan2022basicvsr++}, RVRT \cite{liang2022recurrent}, and our UniWRV.}
	\label{figure_5}
\end{figure}

The quantitative comparison of NIQE \cite{mittal2012making} and BRISQUE \cite{mittal2012no} on real-world adverse weather videos are presented in Table \ref{table_realq}. It can be observed that our method exhibits robust generalization capabilities and restores more realistic and reliable videos, achieving higher perceptual quality. Furthermore, Fig. \ref{figure_5} exhibits real-world video adverse weather removal results driven by BasicVSR++ \cite{chan2022basicvsr++}, RVRT \cite{liang2022recurrent}, and our UniWRV. It is observed that all methods can cope with real-world hybrid weather scenarios while our approach restores more clean and photorealistic results, which finely demonstrate both the realistic reliability of
HWVideo and the robustness and generalization capacity of UniWRV. Meanwhile, this experiment also reveals that real-world scenarios often suffer from multiple superimposed degradations rather than simple one corruption.

\begin{figure}[t]
	\centering
	\includegraphics[width=\linewidth]{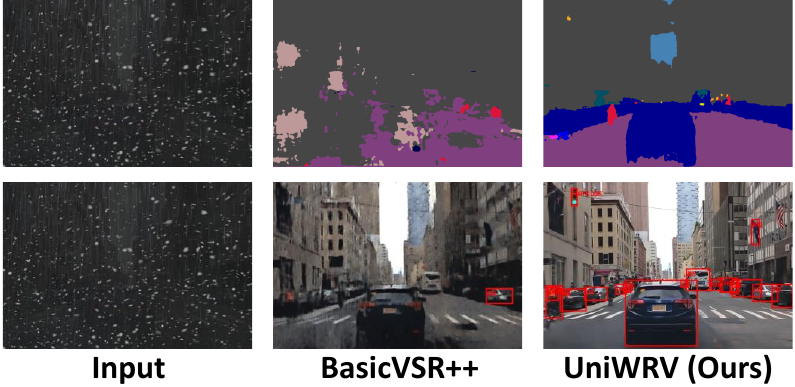}
	\vspace{-17pt}
	\caption{Visual comparisons of segmentation and detection delivered by BasicVSR++ \cite{chan2022basicvsr++} and our UniWRV.}
	\label{figure_6}
\end{figure}

Additionally, we further evaluate the
effect of the restoration results on segmentation and detection with pre-trained DeepLabv3+ \cite{chen2018encoder} and FasterRCNN \cite{ren2015faster}. 
As can be seen from Fig. \ref{figure_6}, the semantics and objects of our restored results can be more accurately identified by pre-trained detection and segmentation models, indicating that our method can restore more reasonably clean details and content.

\begin{table}[h]	
	\footnotesize
	\caption{Quantitative comparison on real-world adverse weather conditions.}
	\tabcolsep=15pt
	\vspace{-5pt}
	\begin{center}
		\begin{tabular}{l|cc}
			\toprule[1pt]
			Method&NIQE $\downarrow$&BRISQUE $\downarrow$\\
			
			\midrule
			Degraded Input Videos & 6.247 &37.36 \\
			EDVR \cite{wang2019edvr}& 5.165 &33.25 \\
			BasicVSR \cite{chan2021basicvsr}& 5.014 &30.57 \\
			BasicVSR++ \cite{chan2022basicvsr++}& 5.111 &32.69\\
			VRT\cite{liang2022vrt} & 4.756 &29.77\\
			Shift-Net \cite{li2023simple}& 4.803 &29.69\\
			RVRT \cite{liang2022recurrent}& 4.899 &29.36 \\
			UniWRV (Ours) & \textbf{4.375}&\textbf{24.42 }\\
			\bottomrule[1pt]
		\end{tabular}
	\end{center}
	\label{table_realq}
	\vspace{-10pt}
\end{table}

\subsection{Generic Unified Model for Various Video Degradations}
Apart from adverse weather removal tasks, we also conduct experiments on the all-in-one noise, blur, and moiré removal from video. Specifically, we first merge the training sets of Set8 \cite{tassano2019dvdnet}, DVD \cite{su2017deep}, and VDD \cite{dai2022video}, and train models on the hybrid training set and then test them on each test set. The quantitative comparisons are shown in Table \ref{table8}. It is observed that our method achieves state-of-the-art results in multiple heterogeneous degradations unified learning, outperforming existing methods by a large margin. This result also exposes the superiority and generalization of our method in unified video restoration with various heterogeneous degradations beyond weather removal.

\begin{table*}[h]
	\setlength{\tabcolsep}{15pt}	
	\caption{Quantitative comparison with the SOTA methods on unified video denoising, video deblurring, and video demoiréing.}
	\footnotesize
	\vspace{-6pt}
	\begin{center}
		\begin{tabular}{c|cccccccc}
			\toprule[1pt]
			\makecell[l]{\multirow{2}{*}{Method}}&\multicolumn{2}{c}{Set8 \cite{tassano2019dvdnet} ($\sigma=30$)} &\multicolumn{2}{c}{Set8 \cite{tassano2019dvdnet} ($\sigma=50$)}&\multicolumn{2}{c}{DVD \cite{su2017deep}}&\multicolumn{2}{c}{VDD \cite{dai2022video}}
			\\
			& PSNR & SSIM&PSNR & SSIM& PSNR & SSIM&PSNR & SSIM\\
			\midrule
			\makecell[l]{EDVR \cite{wang2019edvr}}&29.11&0.8346 &27.12&0.7889&29.15&0.8669&18.02&0.6089\\
			\makecell[l]{BasicVSR \cite{chan2021basicvsr}}&29.21&0.8379 &26.30&0.7996&29.32&0.8778&18.32&0.6415\\
			\makecell[l]{BasicVSR++ \cite{chan2022basicvsr++}}&29.35&0.8498 &27.41&0.8078&29.44&0.8847&18.65&0.6589\\
			
			\makecell[l]{VRT \cite{liang2022vrt}}&29.56&0.8385 &27.47&0.8077&29.51&0.8898&18.87&0.6675\\
			\makecell[l]{Shift-Net \cite{li2023simple}}&29.49&0.8266 &27.51&0.8067&29.53&0.8888&18.93&0.6690\\			
			\makecell[l]{RVRT \cite{liang2022recurrent}}&29.58&0.8379 &27.55&0.8125&28.66&0.8936&19.05&0.6738\\
			\makecell[l]{UniWRV} &\textbf{32.15} &\textbf{0.8818}&\textbf{30.10} &\textbf{0.8611}&\textbf{32.78}&\textbf{0.9365}&\textbf{21.78}&\textbf{0.7236}\\
			\bottomrule[1pt]
		\end{tabular} 
	\end{center}
	\label{table8}
	\vspace{-9pt}
\end{table*}

\subsection{Ablation Studies}\label{abla}

\begin{figure}[h]
	\centering
	\includegraphics[width=.9\linewidth]{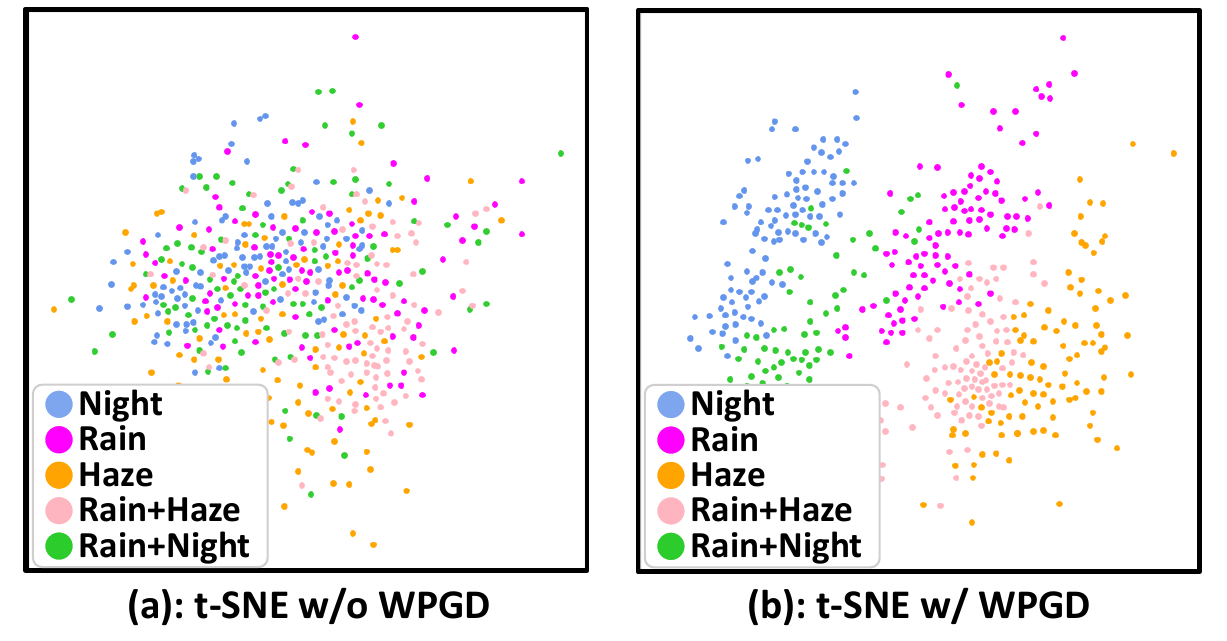}
	\vspace{-6pt}
	\caption{t-SNE visualization of extracted features without and with WPGM.}
	\label{figure_tsne}
	\vspace{-4pt}
\end{figure}

\begin{figure}[h]
	\centering
	\includegraphics[width=.9\linewidth]{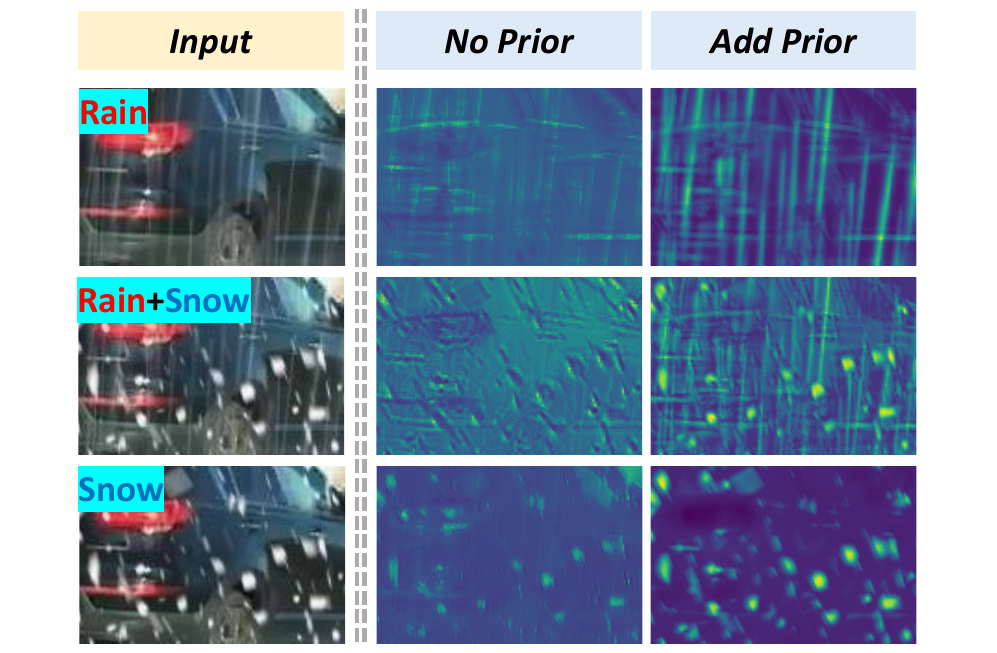}
	\vspace{-6pt}
	\caption{ Feature visualization of three heterogeneous weather
		conditions (i.e., rain, snow, and rain+snow).}
	\label{figure_pfv}
	\vspace{-4pt}
\end{figure}
\noindent\textbf{Effect of weather prior guided module.} During the training process, the weather prior and the subsequent feature extraction module are optimized together, the feature extraction module is responsible for learning the prompts embedded in the prior and incorporating them into the extraction process of the input video feature. We conduct experiments to demonstrate the validity of WPGD in Fig. \ref{figure_tsne}. It is observed that the features extracted after WPGM are clustered into different clusters for different degradation types whereas features of different degradation types are mixed together without WPGM deployment. Furthermore, we have also conducted experiments in Fig. \ref{figure_pfv} to visualize the extracted features of prior addition. It is observed that adding prior allows the feature extraction module to accurately extract specific types of degraded features in the frame whereas without prior it it fails to focus on the degraded features useful for restoration. This phenomenon should be attributed to that WPGM adaptively queries the tailored weather prior for different instances and the feature extraction operation can utilize the queried prior as a prompt to identify the characteristics of the pending features and perform dedicated feature processing.

\begin{figure*}[h]
	\centering
	\includegraphics[width=\linewidth]{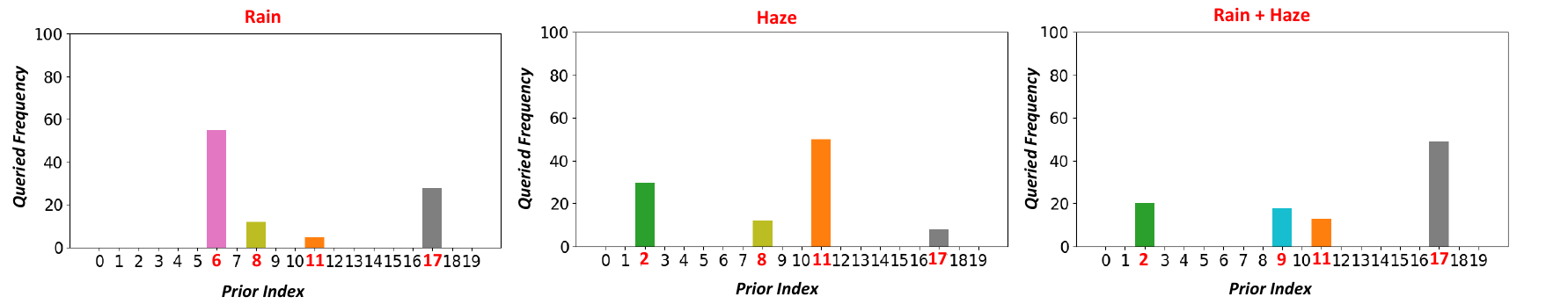}
	\vspace{-17pt}
	\caption{Statistical results of prior index counted from 100 samples for each weather type. Specific prior vectors are queried by specific weather conditions with similar degradation properties.}
	\label{figure_priorindex}
	\vspace{-10pt}
\end{figure*}

Additionally, to more clearly illustrate the correspondence between weather conditions and learned weather priors, we randomly select 100 samples from each of the three weather types Rain, Haze, and Rain+Haze, and count the vector indexes of their queried priors in the fifth layer's weather prior bank. The bar chart of the statistical results is provided in Fig. \ref{figure_priorindex}. It is observed that samples of the same weather type can query multiple specific priors, indicating that samples of the same weather type also have diverse degradation properties, and our method can adaptively learn comprehensive and diverse priors to more accurately characterize diverse degradation properties. Meanwhile, the results also reflect that samples from different weather types may have similar degradation properties.
For example, dense rain streaks exhibit the characteristics of haze, thus some of the "Rain" samples are queried to the "Rain+Haze" prior that index=17. The experiments should demonstrate the consistency between the learned weather prior and specific weather conditions with similar degradation properties, thus the network can identify the degradation properties of pending features by querying corresponding weather priors and using the queried prior as prompts to extract dedicated degradation features for specific scenarios.

\noindent\textbf{Universal weather prior guided module.} As mentioned above, our WPGD can be integrated into existing image restoration architectures boosting their performance on all-in-one multiple weather removal tasks. We deployed WPGD on MPRNet \cite{zamir2021multi}, Restormer \cite{zamir2022restormer}, and NAFNet \cite{chen2022simple} and compared them with state-of-the-art all-in-one methods \cite{valanarasu2022transweather,li2022all,chen2022learning,potlapalli2023promptir}. They are all trained and tested on a combined dataset consisting of
Rain1200 \cite{zhang2018density}, Raindrop \cite{qian2018attentive}, and SOTS \cite{li2018benchmarking}. It can be observed from Table \ref{table3} that the methods that deploy our WPGD achieve new state-of-the-art performance which provides strong evidence of the generality, universality, and effectiveness of our proposed WPGD for multiple heterogeneous weather representation.

\begin{table}
	\centering
	\footnotesize
	\makeatletter\def\@captype{table}\makeatother
	\caption{Results of applying WPGD into image restoration methods. PSNR gains compared to pure training are provided in parenthesis. }
	\label{table3}
	\vspace{-3pt}
	\tabcolsep=28pt
	\begin{tabular}{cc}
		\toprule[1pt]
		Method& PSNR\\
		\midrule
		TransWeather \cite{valanarasu2022transweather} &32.87\\
		AirNet \cite{li2022all}&32.98\\
		TKL \cite{chen2022learning}&33.01\\
		PromptIR \cite{potlapalli2023promptir} &33.10\\
		\midrule
		MPRNet \cite{zamir2021multi} + WPGD  &33.11 ($\uparrow$ 0.41) \\
		Restormer \cite{zamir2022restormer} + WPGD  &33.16 ($\uparrow$ 0.39)\\
		NAFNet \cite{chen2022simple} + WPGD &33.25 ($\uparrow$ 0.45)\\
		\bottomrule[1pt]
	\end{tabular}
\end{table}

\noindent\textbf{Effect of dynamic routing aggregation.} In Table \ref{table_smr}, we compare the complexity and performance of different aggregation modules. Compared to static aggregation, vanilla dynamic routing through adaptive fusion can more comfortably cope with different weather videos bringing a performance improvement of 1.36 dB PSNR, but introduces a high number of parameters and computational complexity. By routing the parameters first and then performing a single computation, the computational complexity of the parameter routing scheme is greatly reduced, but the number of parameters is still high. As a comparison, although our proposed routing strategy shows a slight performance degradation over the vanilla routing scheme, the complexity is much less than it, bringing only 0.1 M parameters and 2 G of FLOPs. 
\begin{table}[h]	
	\footnotesize
	\caption{Comparison of params and FLOPs among different aggregation schemes.}
	\tabcolsep=4pt
	\vspace{-8pt}
	\begin{center}
		\begin{tabular}{lccc}
			\toprule
			Variant& Params (M) &FLOPs (G)&PSNR\\
			
			\midrule
			\makecell[l]{UniWRV w/ static aggregation}  &17.8&161&23.28\\
			\makecell[l]{UniWRV w/ vanilla routing scheme}  &82.98&719&24.65\\
			\makecell[l]{UniWRV w/ parameter routing scheme}  &82.98&162&24.51\\
			\makecell[l]{\textbf{UniWRV (Ours)} } &  \textbf{17.9}&\textbf{162}& \textbf{24.37}\\
			
			\bottomrule
		\end{tabular}
	\end{center}
	\label{table_smr}
	\vspace{-10pt}
\end{table} 

\begin{figure}[h]
	\centering
	\includegraphics[width=\linewidth]{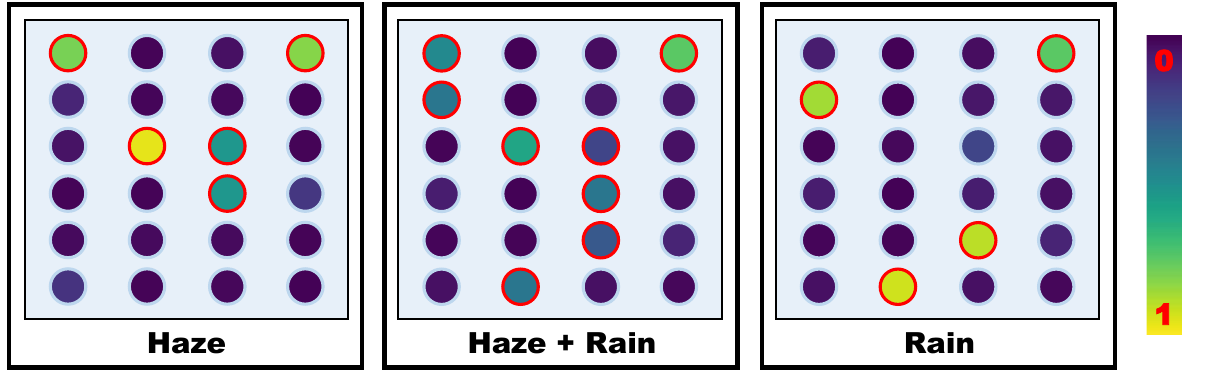}
	\vspace{-12pt}
	\caption{Average routing visualization of different input weather types. Nodes with higher weight are marked by red boxes. }
	\label{figure_routing}
	\vspace{-7pt}
\end{figure}

We have further provided an average routing visualization  (averaged from 100 test frames for each weather type) of three weather types, i.e., haze, rain, and haze+rain) in Fig. \ref{figure_routing}. It is observed that the network dynamically adjusts the routing weights according to the input weather degradation to adaptively fuse adjacent frames of different weather videos. Meanwhile, only specific nodes of the same weather type are assigned with high weights, indicating that samples of the same weather type consistently activate these specific nodes. Furthermore, the paths of the hybrid scenario (haze+rain condition) are a combination of the corresponding single weather paths (haze condition and rain condition), demonstrating that hybrid scenarios also have consistent routing paths with corresponding single weather scenarios.

\begin{figure}[h]
	\centering
	\includegraphics[width=\linewidth]{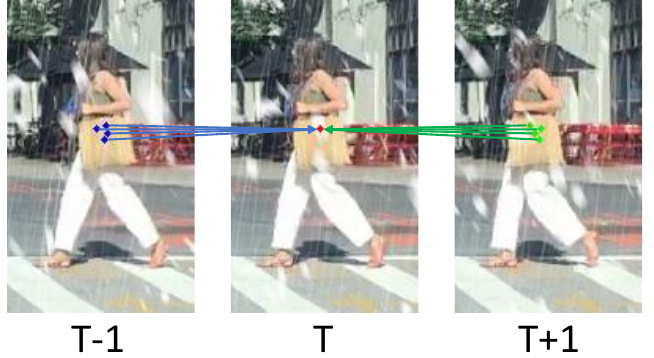}
	\vspace{-10pt}
	\caption{The visualization of deformable multi-frame attention within adjacent frames, which captures available clean features from adjacent frames through deformable sampling points.  }
	\label{figure_def}
	\vspace{-3pt}
\end{figure}

\noindent\textbf{Effect of deformable multi-frame attention.} Considering that different frames are affected by different areas and degrees of weather, deformable multi-frame attention aims to adaptively generate sampling points for neighboring frames as well as the weights of the sampling points to efficiently fuse the valuable features of neighboring frames. We have provided a visualization of the deformable sampling points of adjacent frames in Fig. \ref{figure_def}. It is observed that deformable attention can explore useful content features from clean pixel regions of adjacent frames by dynamically predicting the locations of sampling points.

\noindent\textbf{Universal video weather removal framework.}
As stated above, the feature extraction operation can be substituted with any existing mature block. In this work, we only deploy the simplest residual block to serve as a baseline network and better feature extraction blocks inevitably lead to better performance. The experimental results of substituting for different network blocks are provided in Table \ref{table_uf}. It is observed that the performance of our method improves further when replaced with more advanced network blocks.
\begin{table}[h]	
	\footnotesize
	\caption{Quantitative comparison of different feature extraction blocks.}
	\tabcolsep=15pt
	\vspace{-5pt}
	\begin{center}
		\begin{tabular}{l|cc}
			\toprule[1pt]
			Blocks&PSNR&SSIM\\
			
			\midrule
			UniWRV + Residual Block (Ours)	&24.37&0.8386\\
			UniWRV + NAFNet Block [3]	&24.58&0.8423\\
			UniWRV + Uformer Block [4]	&24.55&0.8411\\
			UniWRV + Restormer Block [5]	&24.64&0.8431\\
			
			\bottomrule[1pt]
		\end{tabular}
	\end{center}
	\label{table_uf}
	\vspace{-10pt}
\end{table}

\subsection{Performance vs. complexity.} As it can be seen from Tab. \ref{table_pc}, our UniWRV enables the unified removal of multiple heterogeneous degradations with moderate complexity. In contrast, existing methods with considerable complexity are primarily tailored for task-specific purposes, and when confronted with the challenge of addressing multiple heterogeneous degradations, they demonstrate a notable decline in performance. Our approach can handle a processing frame rate of 2 FPS on a single 3090 graphics card, while processing real-time video at a frame rate of 30 FPS requires about 500 TFLOPS of arithmetic power. With the current hardware support, our approach couldn't be applied in real time but it is an important step forward. Because existing methods are comparable to our inference speed but can only cope with one weather scenario while our method can cope with multiple scenarios, which greatly improves the deployment efficiency.

\begin{table}[h]	
	\footnotesize
	\caption{Comparison of params and runtime among SOTA adverse weather removal methods.}
	\tabcolsep=10pt
	\vspace{-5pt}
	\begin{center}
		\begin{tabular}{lccc}
			\toprule[1pt]
			Method& Params (M) &Runtime (ms)&PSNR\\
			
			\midrule
			\makecell[l]{EDVR \cite{wang2019edvr}}&20.6&1245&19.21\\
			\makecell[l]{BasicVSR \cite{chan2021basicvsr}}&6.3 &375&19.77\\
			\makecell[l]{BasicVSR++ \cite{chan2022basicvsr++}}& 7.3&428&20.29\\
			\makecell[l]{VRT \cite{liang2022vrt}} &18.3&815&20.79\\
			\makecell[l]{Shift-Net \cite{li2023simple}} &12.9&327&20.89\\
			\makecell[l]{RVRT \cite{liang2022recurrent}}	&13.6&208&21.06\\
			\makecell[l]{UniWRV } &  \textbf{19.9}&\textbf{454}& \textbf{24.37}\\
			
			\bottomrule[1pt]
		\end{tabular}
	\end{center}
	\label{table_pc}
	\vspace{-15pt}
\end{table} 

\section{Limitations and Future Work} \label{g}

\begin{figure}[h]
	\centering
	\includegraphics[width=\linewidth]{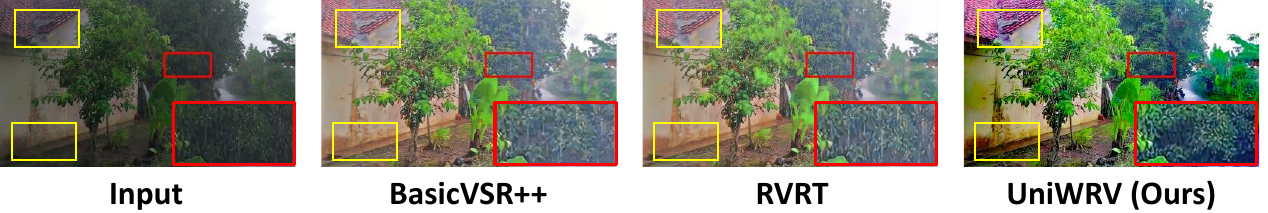}
	\vspace{-10pt}
	\caption{One failure case of out method on real-world hybrid adverse weather conditions.  }
	\label{figure_fa}
	\vspace{-9pt}
\end{figure}

One failure case of our method is depicted in Fig. \ref{figure_fa}. It is observed that although our method removes rain and haze occlusions to reconstruct high-quality details, the roof, and the exterior walls of buildings are over-restored, which decreases the realism of the image. This is mainly attributed to that it suffers from combined damage of rain, haze, and low-light, with background details heavily obscured and our method over-restored some specific areas in order to recover higher quality results without finely balancing the visual effect of the restoration with the realism. In contrast, the compared approach restored with insufficient overall force, and although no over-restore was present but left visible rain and haze degradation and low-contrast results. Going forward, we are committed to further optimizing our method to ensure that our method can gracefully restore failure cases with a fine balance between enhancing degraded images and preserving realism. 
		Another potential limitation of our UniWRV, which is shared
with the existing image restoration algorithms, is the reliance on large-scale paired data. As stated in the main paper, there are $900K$ pairs of data for model training. On the other hand, UniWRV can handle arbitrary hybrid conditions combined with four common weather, while other rare adverse weather like frost on glass, sand, and dust has remained to be addressed. In the future, we will focus on training models that can handle complicated hybrid conditions with single degraded data, which is extremely challenging but meaningful work, especially when more weather types are taken into account, whose number of hybrid conditions grows exponentially. Additionally, there is an inevitable domain gap between synthetic and real-world videos, how to capture and build a real paired dataset is also a very worthwhile research issue, especially for improving the generalization of the model.

\section{Concluding Remarks}\label{5}

In this work, we proposed a novel unified model, namely UniWRV, to remove multiple hybrid adverse weather from video in an all-in-one fashion, which exhibits a remarkable ability to automatically adapt to uncertain heterogeneous video degradation distributions. In contrast to existing frameworks that deal with different weather conditions individually and rely on weather type identification to
indicate which task to perform,
UniWRV can remove uncertain hybrid adverse weather with a single trained model in one go, which is more competitive and practical to a variety of real-world scenarios wherein multiple weather occur simultaneously and change with time and space.
In addition, we propose a new hybrid adverse weather video dataset HWVideo to simulate real-world multiple hybrid weather scenarios, which contains 1500 adverse-weather/clean paired video clips of 15 weather conditions. Extensive experiments manifest the effectiveness, superiority, and robustness of our proposed UniWRV as well as the comprehensiveness and authenticity of HWVideo. Additionally, our proposed UniWRV empowers robust heterogeneous degradation learning capability in various generic video restoration tasks beyond weather degradations.
We expect this work to provide
insights into multiple heterogeneous video degradation learning
and steer future research on this Gordian knot.

\bibliographystyle{IEEEtran}
\bibliography{reference}

\end{document}